\documentclass[runningheads]{llncs}
\usepackage{graphicx}
\usepackage{times}
\usepackage{epsfig}
\usepackage{graphicx}
\usepackage{amsmath}
\usepackage{amssymb}
\usepackage{wrapfig,lipsum,booktabs}
\usepackage{times}  
\usepackage{helvet} 
\usepackage{courier}  
\usepackage[hyphens]{url}  
\usepackage{graphicx} 
\usepackage{subfig}
\usepackage{amsmath}
\usepackage{multirow}
\usepackage[ruled,vlined]{algorithm2e}
\usepackage{caption}
 \usepackage[misc]{ifsym}

\usepackage{caption} 

\makeatletter
\newcommand{\removelatexerror}{\let\@latex@error\@gobble}
\makeatother
\begin{document}
\title{Deep Model Compression Via Two-Stage Deep Reinforcement Learning}
\toctitle{Deep Model Compression Via Two-Stage Deep Reinforcement Learning}
%
%
\author{Huixin Zhan\inst{1}\orcidID{0000-0001-8926-1941} \and
Wei-Ming Lin\inst{2}\orcidID{0000-0002-9350-6646} \and
Yongcan Cao{\Letter}\inst{2}\orcidID{0000-0003-3383-0185}}

\tocauthor{Huixin Zhan, Wei-Ming Lin, and Yongcan Cao}
\authorrunning{Zhan et al.}

%
\institute{Texas Tech University, Lubbock TX 79415, USA \\
\email{huixin.zhan@ttu.edu} \and
The University of Texas at San Antonio, San Antonio TX 78249, USA\\
\email{\{weiming.lin,yongcan.cao\}@utsa.edu}}

\maketitle              
\begin{abstract}
Besides accuracy, the model size of convolutional neural networks (CNN) models is another important factor considering limited hardware resources in practical applications. For example, employing deep neural networks on mobile systems requires the design of accurate yet fast CNN for low latency in classification and object detection. To fulfill the need, we aim at obtaining CNN models with both high testing accuracy and small size to address resource constraints in many embedded devices. In particular, this paper focuses on proposing a generic reinforcement learning-based model compression approach in a two-stage compression pipeline: pruning and quantization. The first stage of compression, i.e., pruning, is achieved via exploiting deep reinforcement learning (DRL) to co-learn the accuracy and the FLOPs updated after layer-wise channel pruning and element-wise variational pruning via information dropout. The second stage, i.e., quantization, is achieved via a similar DRL approach but focuses on obtaining the optimal bits representation for individual layers. We further conduct experimental results on CIFAR-10 and ImageNet datasets. For the CIFAR-10 dataset, the proposed method can reduce the size of VGGNet by $9\times$ from $20.04$MB to $2.2$MB with a slight accuracy increase.
For the ImageNet dataset, the proposed method can reduce the size of VGG-16 by $33\times$ from $138$MB to $4.14$MB with no accuracy loss.
\end{abstract}

\section{Introduction}

CNN has shown advantages in producing highly accurate classification in various computer vision tasks evidenced by the development of numerous techniques, e.g., VGG~\cite{simonyan2014very}, ResNet~\cite{he2016deep}, DenseNet~\cite{huang2017densely}, and numerous automatic neural architecture search approaches~\cite{veit2018convolutional,yu2018learning}. Albeit promising, the complex structure and large number of weights in these neural networks often lead to explosive computation complexity. Real world tasks often aim at obtaining high accuracy under limited computational resources. This motivates a series of works towards a light-weight architecture design and better speed-up ratio-accuracy trade-off, including Xception~\cite{chollet2017xception}, MobileNet/MobileNet-V2~\cite{howard2017mobilenets}, ShuffleNet~\cite{zhang2018shufflenet}, and CondenseNet~\cite{huang2018condensenet}, where group and deep convolutions are crucial.

In addition to the development of the aforementioned efficient CNN models for fast inference, many results have been reported on the compression of large scale models, e.g., reducing the size of large-scale CNN models with little or no impact on their accuracies. Examples of the developed methods include low-rank approximation~\cite{denton2014exploiting,lebedev2014speeding}, network quantization~\cite{rastegari2016xnor,wang2019haq}, knowledge distillation~\cite{hinton2015distilling}, and weight pruning~\cite{han2015deep,zhuang2018discrimination,he2017channel,jia2018droppruning,liu2017learning}, which focus on identifying unimportant channels that
can be pruned. However, one key limitation in these methods is the lack of automatic learning of the pruning policies or quantization strategies for reduced models.

Instead of identifying insignificant channels and then conducting compression during training, another potential approach is to use reinforcement learning (RL) based policies to determine the compression policy automatically. There are limited results on RL based model compression~\cite{he2018amc,wu2018pocketflow}. In particular,~\cite{he2018amc,wu2018pocketflow} proposed a deep deterministic policy gradient
(DDPG) approach that uses reinforcement learning to efficiently sample the designed space for the improvement of model compression quality. While DDPG can provide good performance in some cases, it often suffers from performance volatility with respect to the hyper-parameter setup and other tuning methods. Besides, these RL-based methods don't directly deal with leveraging the sparse features of CNN, i.e., pruning the small weight connections. 

Recently, RL based search strategies have been developed to formulate neural architecture search. For example,~\cite{zhong2018practical,zoph2018learning} considered the generation of a neural architecture via considering agent’s action space as the search space in order
to model neural architecture search as a RL problem. Different RL approaches were developed to emphasize different representations of the agent’s policies along with the optimization methods. In particular,~\cite{zoph2018learning} used a recurrent neural network based policy to sequentially sample a string that in turn encodes the neural architecture. Both REINFORCE policy gradient algorithm~\cite{sutton2000policy} and Proximal Policy Optimization (PPO)~\cite{schulman2017proximal} were used to train the network. Differently,~\cite{baker2016designing} used Q-learning to train a policy that sequentially chooses the type of each layer and its corresponding hyper-parameters. Note that~\cite{zhong2018practical,zoph2018learning} focuses on generating CNN models with efficient architectures, while not on the compression of large scale CNN models.

In this paper, we propose to develop a novel two-stage DRL framework for deep model compression. In particular, the proposed framework integrates layer-wise pruning rate learning based on testing accuracy and FLOPs, element-wise variational pruning, and per-layer bits representation learning. In the pruning stage, we first conduct channel pruning that will prune the input channel dimension (i.e., C dimension) with minimized accumulated error in feature maps with the obtained per-layer pruning rate. Then fine-tuning with element-wise pruning via information dropout is conducted to prune the weights in the kernel (i.e., from H and W dimensions).

Briefly, this paper has three main contributions:

1. We propose a novel DRL algorithm that can obtain stabilized
policy and address Q-value overestimation in DDPG
by introducing four improvements: (1) computational constrained
PPO: Instead of collecting $T$ timesteps of action
advantages in each of $M$ parallel actors and updating the
gradient in each iteration based on $MT$ action advantages in
one iteration of the typical PPO, we propose to collect Q-values in each tilmestep of $M$ parallel actors and update the gradient each timestep based on the $M$ sampled Q-values;
(2) PPO-Clip Objective:
We propose to modify the expected return of the policy by clipping subject to policy change penalization.
(3) smoothed policy update: Our algorithm first enables
multiple agents to collect one minibatch of Q-values based on
the prior policy and updates the policy while penalizing policy
change. The target networks are then updated by slowly
tracking the learned policy network and critic network; and
(4) target policy regularization: We propose to smooth Q-functions
along regularized actions via adding noise to the
target action. The four improvements altogether can substantially
improve performance of DRL to yield more stabilized
layer-wise prune ratio and bit representations for deep compression,
hence outperforming the traditional DDPG. We experimentally
show the volatility of DDPG-based compression method in order to backup some common failure mode of policy exploitation in DDPG-based method as shown in Figure~\ref{fig:graph}.

2. Pruning: We propose a new \textbf{ppo} with variational pruning compression structure with element-wise variational pruning that can prune three dimensions of CNN. We further learn the \textit{Pareto front} of a
set of models with two-dimensional outputs, namely, model
size and accuracy, such that at least one output is better than,
or at least as good as, all other models by constraining the
actions. More compressed models can be obtained with little
or no accuracy loss.

3. Quantization: We propose a new quantization method that
uses the same DRL-supported compression structure, where
the optimal bit allocation strategy (layer-wise bits representation)
is obtained in each iteration via learning the updated accuracy.
Fine-tuning is further executed after each rollout.

\section{A Deep Reinforcement Learning Compression Framework}\label{sec:drl}
\begin{figure}
\centering
\includegraphics[width=0.9\columnwidth]{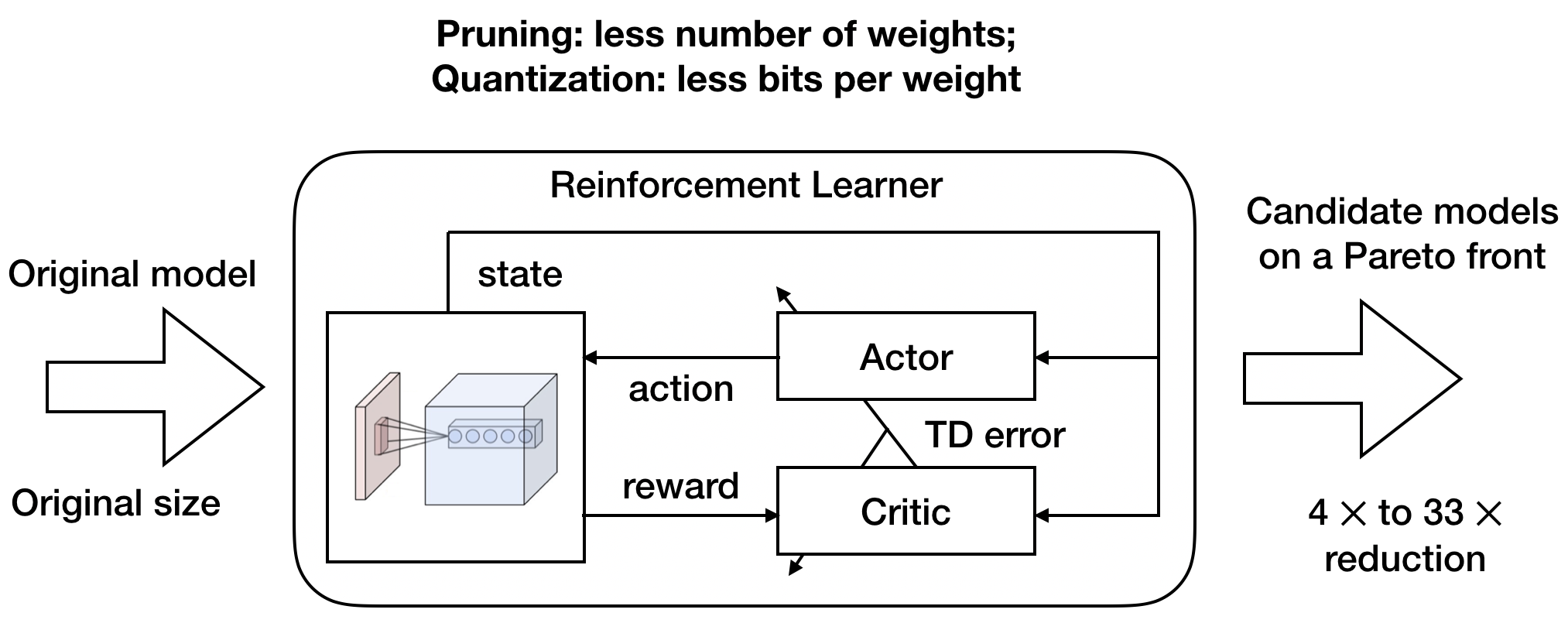} 
\caption{The proposed deep reinforcement learning compression framework.}
\label{fig1}
\end{figure}

\begin{figure*}  
\centering
\subfloat[MobileNet-v1 on ILSVRC-2012.]{\includegraphics[width=5.cm,height=3.cm]{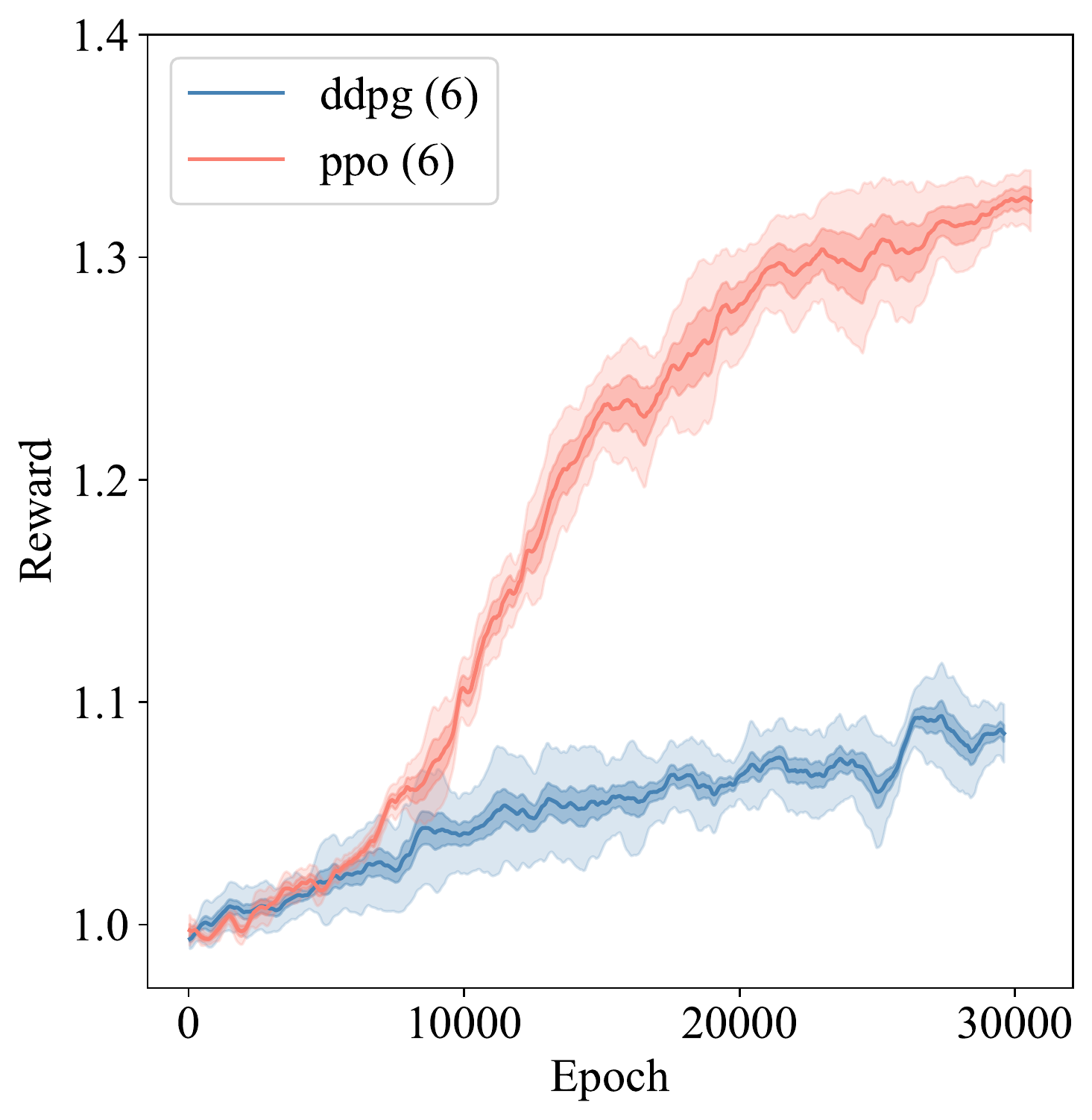}\label{rr1}}\hspace{15pt}
\subfloat[MobileNet-v2 on ILSVRC-2012.]{\includegraphics[width=5.cm,height=3.cm]{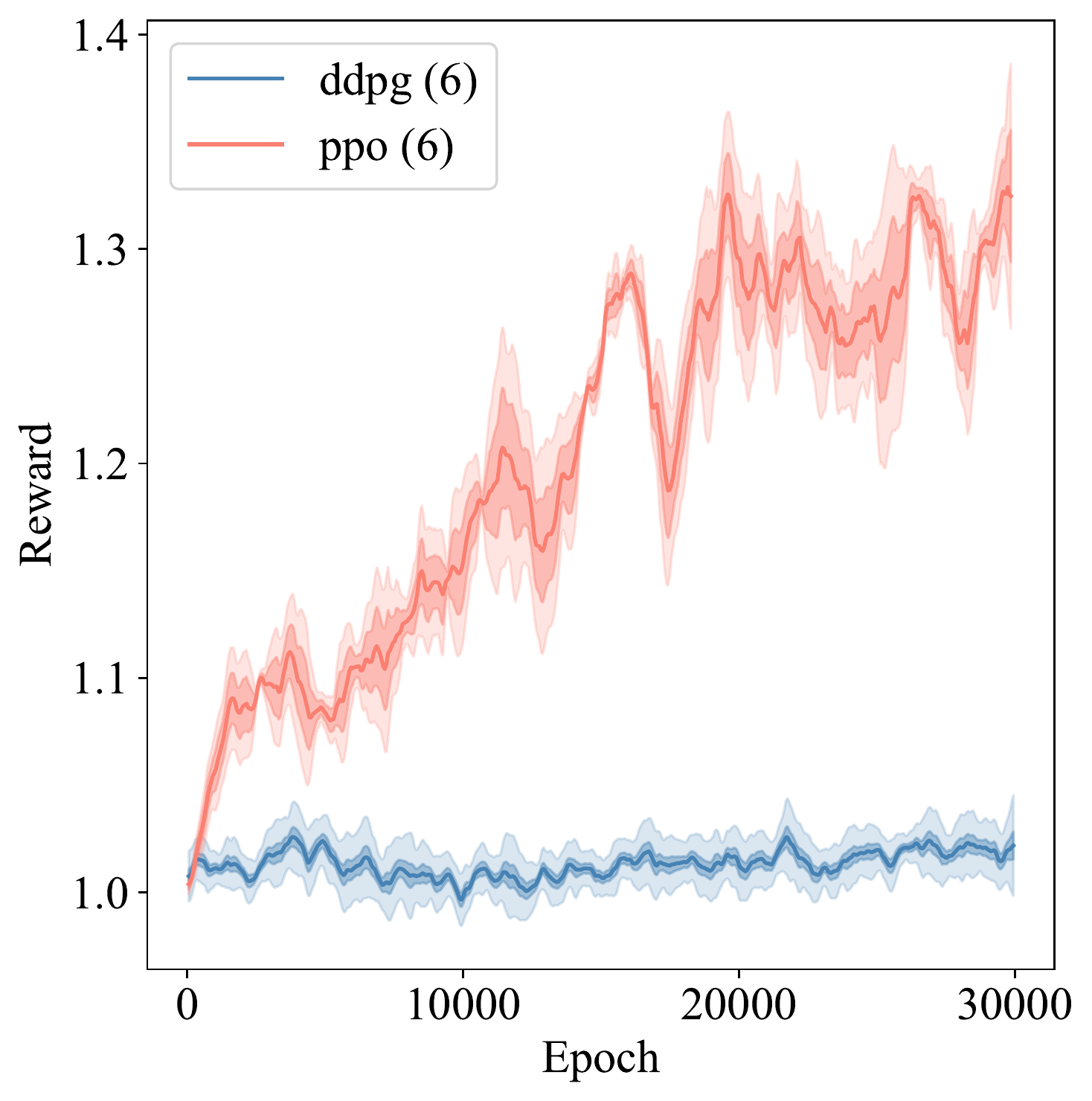}\label{rr2}}
\caption{Comparison of RL-based pruning methods, e.g., \textbf{ppo (modified)} and \textbf{ddpg}, for MobileNet-v1 and MobileNet-v2 on ILSVRC-2012 for $6$ runs.}
\label{fig:graph}
\end{figure*}

In this section, we focus on presenting the proposed new generic reinforcement learning based model compression approach in a two-stage compression pipeline: pruning and quantization. Figure~\ref{fig1} shows the overall structure.
The two-stage compression pipeline includes pruning and quantization. Adopting the pipeline can achieve a typical model compression rate between $4 \times$ and $33 \times$. Investigating the Pareto front of candidate compression models shows little or no accuracy loss.

\subsection{State}
In both pruning and quantization, in order to discriminate each layer in the neural network, we use a $8$-dimension vector space to model a continuous state space:
\begin{equation}
s_t = [N_{Lr}, N, C, H, W, \textit{Stride}, A_H^t, \textit{FLOPs}],
\end{equation}
where $N_{Lr}$ is the index of the layer, $N$ and $C$ are the dimension of, respectively, output channels and input channels, $H$ is the kernel height, $W$ is the kernel width, $\textit{Stride}$ is the number of pixels shifts over the input matrix, $A_H^t$ is the maximum pruning rate in pruning (respectively, the maximum and minimum bits representation in quantization) with respect to layer $t$, and $\textit{FLOPs}$ is the number of floating point operations in each layer.

\subsection{Action}
In pruning, determining the compression policy is challenging because the pruning rate of each layer in CNN is related in an unknown way to the accuracy of the post-compression model. Since our goal is to simultaneously prune the $C$, $H$, and $W$ dimensions. As the dimension of channels increases or the network goes deeper, the computation complexity increases exponentially. Instead of searching over a discrete space, a continuous reinforcement learning control strategy is needed to get a more stabilized scalar continuous action space, which can be represented as $a_t = \left\{pr_t | pr_t \in [pr_h, pr_l]\right\}$, where $pr_l$ and $pr_h$ are the lowest and highest and pruning rates, respectively. The compression rate in each layer is taken as a replacement of high-dimensional discrete masks at each
weight of the kernels. Similarly, in quantization, the action is also modeled in a scalar continuous action space, which can be represented as $a_t = \left\{b_t|b_t \in \mathbb{N}^+\right\}$, where $b_t$ is the
number of bits representation in layer $t$.

\subsection{Reward}
To evaluate the performance of the proposed two-stage compression pipeline, we propose to construct two reward structures, labeled $r1$ and $r2$. $r1$ is a synthetic reward system as the normalization of current accuracy and FLOPs. $r2$ is an accuracy-concentrated reward system. In pruning, the reward
for each layer can be chosen from $r_t \in \left\{ r1, r2\right\}$. In quantization, we use $r2$ as our selected reward structure. In particular, $r1 = 1- \frac{FLOPs_t - FLOPs_{low}}{FLOPs_{high} - FLOPs_{low}} + p_{ac}$ and $r2 = p_{ac}$ , where $p_{ac}$ is the current accuracy, $FLOPs_{high}$ and $FLOPs_{low}$ are the highest and lowest FLOPs in observation.
\subsection{The Proposed DRL Compression Structure}\label{sec:DRLstructure}
In the proposed model compression method, we learn the Pareto front of a set of models with two-dimensional outputs (model size and accuracy) such that at least one output is better than (or at least as good as) all other outputs. We adopt a popular asynchronous actor critic~\cite{mnih2016asynchronous} RL framework to compress a pre-trained network in each layer sequentially. At time step $t$, we denote the observed state by $s_t$, which corresponds to the per-layer features. The action set is denoted by $\mathcal{A}$ of size $1$. An action, $a_t \in \mathcal{A}$, is drawn from a policy function distribution: $a_t \sim \mu(s_t| \theta^\mu) +\mathcal{N}_t \in \mathbb{R}^1$, referred to as an actor, where $\theta^\mu$ is the current policy network parameter and the noise $\mathcal{N}_t \in \mathcal{N}(0,\epsilon)$. The actor receives the state $s_t$, and outputs an action $a_t$. After this layer is compressed with
pruning rate or bits representation $a_t$, the environment then returns a reward $r_t$ according to the reward function structure $r1$ or $r2$. The updated state $s_{t+1}$ at next time step $t+1$ is observed by a known state transition function $s_{t+1} = f(s_t; a_t)$, governed by the next layer. In this way, we can observe a random minibatch of transitions consisting of a sequence of tuples $\mathbb{B} =\left\{(s_t; a_t; r_t; s_{t+1}) \right\}$. In typical PPO, the surrogate objective is represented by $\hat{\mathbb{E}}_t [\frac{\pi^{\theta^\mu}(a|s_t)}{\pi^{\theta^{\mu^-}}(a|s_t)}\hat{A}_t]$, where the expectation $\hat{\mathbb{E}}_t[\cdot]$ is the empirical average over a finite batch of samples and $\theta^{\mu^-}$ is the prior policy network parameter. If we compute the action advantage $\hat{A}_t$ in each layer, $T$-step time difference rewards are needed, which is computationally intensive. In resource constrained PPO, we propose to replace the action advantages by Q-functions given by $Q(s_t,a_t) = \mathbb{E} [\sum_{i=t}^{t+T} \gamma^{i-t}r_i|s_t,a_t]$, referred to as critic.

The policy network parameterized by $\theta^\mu$ and the value function parameterized by $\theta^Q$ are then jointly modeled by two neural networks. Let $a = \mu (s_i | \theta^\mu)$, we can learn $\theta^Q$ via Q-function regression, namely, Equation~\eqref{pi_theta}, and learn $\theta^\mu$ over the tuples $\mathbb{B}$ with PPO-Clip objective stochastic policy gradient, namely, Equation~\eqref{pi_mu} as
\begin{align}
\theta^Q = \underset{\theta}{\arg\min} \frac{1}{|\mathbb{B}|} \sum_{(s_i,a_i,r_i,s_{i+1}) \in \mathbb{B}}(y_i - Q(s_i,a_i|\theta))^2,\label{pi_theta}
\end{align}
\begin{align}
\theta^\mu = \underset{\theta}{\arg\max} \underset{(s_i,a_i,\cdots) \in \mathbb{B}}{\hat{\mathbb{E}}} \min\left\{  \frac{\pi^{\theta}(a|s_i)}{\pi^{\theta^{\mu^-}}(a|s_i)} Q(s_i,a|\theta^Q),\right.\notag\\ \left.
clip(\frac{\pi^{\theta}(a|s_i)}{\pi^{\theta^{\mu^-}}(a|s_i)},1-c,1+c ) \times Q(s_i,a|\theta^Q) \right\},\label{pi_mu}
\end{align}
where $c$ is the probability ratio of the clipping. A pseudocode of DRL compression structure is shown in Algorithm~\ref{alg:1}.

\scalebox{0.9}{
\begin{minipage}{1.0\linewidth}
\begingroup
\removelatexerror
\begin{algorithm*}[H]\small
\LinesNumbered
\SetAlgoLined
\KwData{Randomly initialize critic network $Q(s, a|\theta^Q)$ and actor $\mu(s|\theta^\mu)$ with weights $\theta^Q$ and $\theta^\mu$. Initialize target network $Q^\prime$ and $\mu^\prime$ with weights $\theta^{Q^\prime} \leftarrow \theta^Q$, $\theta^{\mu^\prime} \leftarrow \theta^\mu$, the learning rate of the target network $\rho$, $A_H^t$, and empty replay buffer $\mathcal{D}$}
\KwResult{Weights $\theta^Q$ and $\theta^\mu$. }
 initialization\;
 \While{Episode $< M$}{
 Initialize a random process $\mathcal{N}$ for action exploration\;
Receive initial observation state $s_1$\;
$M \leftarrow M+1$\;
  \For{$t = 1,\cdots,T$}{
   Select action $a_t = clip(\mu(s_t|\theta^\mu) + \mathcal{N}_t, A_H^t)$ according to the current policy and exploration noise\;
   Execute $a_t$\;
   Store $(s_t,a_t,r_t,s_{t+1})$ in replay buffer $\mathcal{D}$\;
\For{$t=1,\cdots$}{
   Sample a random minibatch of $\mathbb{B}$ trajectories from $\mathcal{D}$\;
   Set $y_t = r_t + \gamma Q(s_{t+1}, \mu(s_{t+1})| \theta^{Q^{\prime}})$\;
    Update the policy by maximizing the “surrogate” objective via stochastic gradient ascent with Adam in Equation~\ref{pi_mu}\;
  Pruning the $C$ dimension of $t$-th layer with pruning rate $a_t$\;
  Executing the element-wise variational pruning in Algorithm~\ref{alg:2}\;
   Update the critic by minimizing the combinatorial loss via stochastic gradient descent in Equation~\ref{pi_theta}\;
   Update the target networks via $\theta^{Q^\prime} \leftarrow  \rho \theta^{Q^\prime} + (1 - \rho) \theta^Q$, $\theta^{\mu^\prime} \leftarrow  \rho \theta^{\mu^\prime} + (1 - \rho) \theta^\mu$\;}
     }
   
 }
 \caption{The proposed DRL compression structure in pruning.}\label{alg:1}
 \end{algorithm*}
 \endgroup
 \end{minipage} 
 }
\section{Pruning}\label{sec:vae1}
In this section, we present two schemes to compress CNN with little or no loss in accuracy by employing reinforcement learning to co-learn the layer-wise pruning rate and the element-wise variational pruning via information dropout. Similar to the aforementioned a3c framework, the layer-wise pruning rate is computed by the actor. After obtaining pruning rate $a_t$, layer $s_t$ is pruned by a typical channel pruning method~\cite{he2017channel}, whose detail will be given below, to select the most representative channels and reduce the accumulated error of feature maps. In other words, after we get the pruning rate, channel pruning can be used to determine which specific channels are less important or we can simply prune based on the weight magnitude. In each iteration, the CNN layer is further compressed by variational pruning. In particular, we start by learning the connectivity via normal network training. Then, we prune the small-weight connections: all connections with weights that create a representation of the data that is minimal sufficient for the task of reconstruction are remained. Finally, we retrain the network to learn the final weights for the remaining sparse connections. 

\begin{figure}
\centering
\includegraphics[width=0.5\columnwidth]{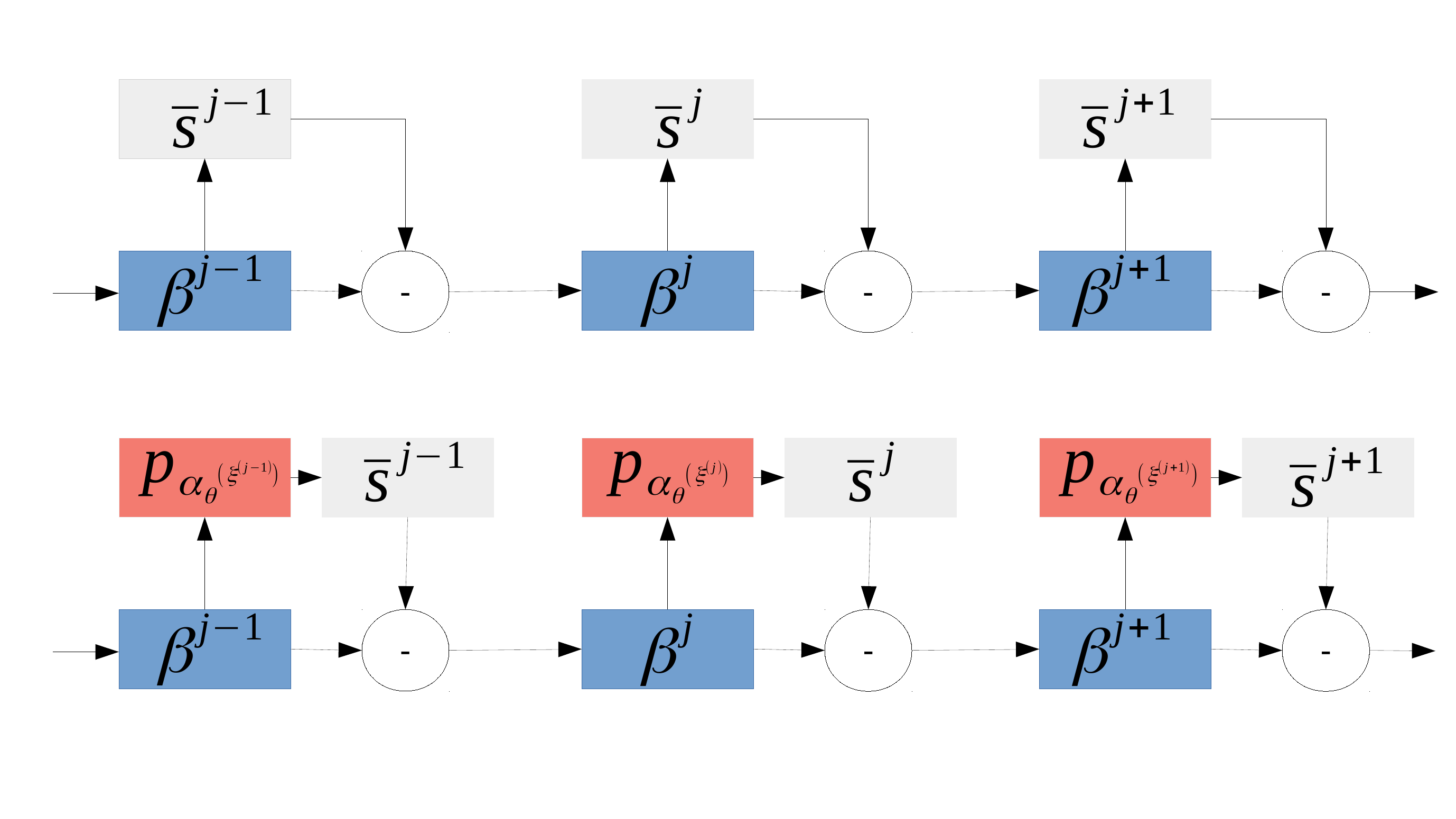} 
\caption{Comparison of the pruning on two aforementioned schemes, i.e., channel pruning and variational pruning via information dropout.}
\label{fig2}
\end{figure}

In pruning, $\beta$ is a vector whose dimension matches the 4D tensor with shape $N \times C \times W \times H$ in each layer. We also define $\beta^i$, the $i$-th entry of $\beta$, as a binary mask for each weight in the kernel. Figure~\ref{fig2} shows the pruning flow in our two schemes. In DRL compression framework, the scalar mask of the $j$-th weight $w_j$ with mask $\beta^j$ is set to zero if the weights are pruned based on LASSO regression, discussed in subsection. If pruned, these weights are moved to $\bar{s}^j$, defined as a set of pruned weights. Otherwise, if the weights are pruned based on information dropout, discussed in subsection, the scalar mask of the $j$-th weight $w_j$ with mask $\beta^j$ will be moved to $\bar{s}^j$ with probability $p_{a_\theta}(\xi^{(j)})$. The weights that play more important role in reducing the classification error are less likely to be pruned.

\subsection{Pruning from $C$ Dimension: Channel Pruning}
The $C$-dimension channel pruning can be formulated as: 
\begin{align}
\underset{\beta,W}{\arg \min} \frac{1}{2N} & \left\|Y - \sum_{i=1}^C \beta X_i W_i^T \right\|_F^2 + \lambda \left\| \beta  \right\|_1\\
\textit{subject~to} &\left\| \beta \right\|_0 \leq pr \times C\notag\\
&\left\| W_i \right\|_F =1, \forall i \notag,
\end{align}
where $pr$ is the pruning rate, $X_i$ and $Y$ are the input volume and the output volume in each layer, $W_i$ is the weights, $\beta$ is the coefficient vector of length $C$ for channel
selection, and $\lambda$ is a positive weight to be selected by users. Then we assign $\beta_i \leftarrow \beta_i \left\| W_i \right\|_F$ and $W_i \leftarrow \frac{W_i}{\left\|W_i\right\|_F}$.

\subsection{Pruning from $H$ and $W$ Dimensions: Variational Pruning}\label{sec:vae11}
In information dropout, we propose a solution to: (1) efficiently approximate posterior inference of the latent variable $\mathbf{z}$ given an observed value $\mathbf{x}$ based on parameter $\theta$, where $\mathbf{z}$ is a representation of $\mathbf{x}$ and defined as some (possibly nondeterministic) function of $\mathbf{x}$ that has some desirable properties in some coding tasks $\mathbf{y}$ and (2) efficiently approximate marginal inference of the variable $\mathbf{x}$ to allow for various inference tasks where a prior over $\mathbf{x}$ is required.

Without loss of generality, let us consider Bayesian analysis of some dataset $\mathcal{D} =\left\{(\mathbf{x}^{(i)},\mathbf{y}^{(i)})\right\}_{i=1}^N$ consisting of $N$ i.i.d samples of some discrete variable $\mathbf{x}$. We assume that the data are generated by some random process, involving an unobserved continuous random variable $\mathbf{z}$. Bayesian inference in such a scenario consists of (1) updating some initial belief over parameters $\mathbf{z}$ in the form of a prior distribution $p_{\theta^\star}(\mathbf{z})$, and (2) a belief update over these parameters in the form of (an approximation to) the posterior distribution $p_\theta(\mathbf{z}| \mathbf{x})$ after observing data $\mathbf{x}$. In variational inference~\cite{kingma2013auto}, inference is considered as an optimization problem where we optimize the parameters $\theta$ of some parameterized model $p_\theta(\mathbf{z})$ such that $p_\theta(\mathbf{z})$ is a close approximation of $p_\theta(\mathbf{z}|\mathbf{x})$ as measured by the KL divergence $D_{KL}(p_\theta(\mathbf{z}|\mathbf{x})| p_\theta(\mathbf{z}))$. The divergence between $p_\theta(\mathbf{z}|\mathbf{x})$ and
the true posterior is minimized by minimizing the negative variational lower bound $\mathcal{L}(\theta)$ of the marginal likelihood of the data, namely,
\begin{align}\label{eq:noise}
\mathcal{L}(\theta, \theta^\star; \mathbf{x}^{(i)}) &= - \sum_{(\mathbf{x}^{(i)},\mathbf{y}^{(i)})\in \mathcal{D}} \mathbb{E}_{p_\theta(\mathbf{z}|\mathbf{x}^{(i)})}[\log p (\mathbf{y}^{(i)}|\mathbf{z})]\notag\\
&+ \alpha D_{KL}(p_\theta(\mathbf{z}|\mathbf{x}^{(i)})| p_\theta^{\star}(\mathbf{z})).
\end{align}

As shown in~\cite{kingma2015variational}, the neural network weight parameters $\theta$ are less likely to overfit the training data if adding input noise during training. We propose to represent $\mathbf{z}$ by computing a deterministic map of activations $f(\mathbf{x})$, and then multiply the result in an element-wise manner by a random noise $\xi$, drawn from a parametric distribution $p_a$ with the variance that depends on the input data $\mathbf{x}$, as
\begin{align}
\mathbf{z} &= (\mathbf{x} \circ \xi) \theta,\notag\\
\xi_{i,j} &\sim p_{a_\theta}(\mathbf{x})(\xi_{i,j}),
\end{align}
where $\circ$ denotes the element product operation of two vectors. A choice for the distribution $p_{a_\theta(\mathbf{x})}(\xi_{i,j})$ is the log-normal distribution $\log(p_{a_\theta(\mathbf{x})}(\xi_{i,j})) = \mathcal{N}(0, a_\theta^2(\mathbf{x}))$~\cite{achille2018information} that makes the normally fixed dropout rates $p_a$ adaptive to the input data, namely,
\begin{align}
\log(p_{a_\theta(\mathbf{x})}(\xi_{i,j})) &\sim \mathcal{N}(\mathbf{z}; 0, a_\theta^2(\mathbf{x})I),\notag\\
\log(p_{\theta^\star}) &\sim \mathcal{N}(\mathbf{z}; \mu, \sigma^2 I),
\end{align}
where $a_\theta(\mathbf{x})$ is an unspecified function of $\mathbf{x}$. The resulting estimator becomes
\begin{align}\label{vad}
\mathcal{L}(\theta; \mathbf{x}^{(i)}) &\sim \frac{1}{N}  \sum_{j = 1}^N [- \log p (\mathbf{y}^{(i)}|\mathbf{z}^{(i,j)})]\notag\\
&+ \alpha [\frac{1}{2 \sigma^2}(a^2_\theta(\mathbf{x}^{(i)})+\mu^2) - \log \frac{a^2_\theta(\mathbf{x}^{(i)})}{\sigma} - \frac{1}{2}],
\end{align}
where $\mathbf{z}^{(i,j)} \sim ( \mathbf{x}^{(i)} \circ \xi^{(i,j)})\theta$ and $\xi^{(i,j)} \sim p_{a_\theta}(\xi) = \log \mathcal{N} (0, a^2_\theta(\mathbf{x}))$. This loss can be optimized using stochastic gradient descent. A pseudocode of this variational pruning is shown in Algorithm~\ref{alg:2} and an illustrative experiment is given in subsection~\ref{sec:vpe}. 

\scalebox{0.9}{
\begin{minipage}{1.0\linewidth}
\begingroup
\removelatexerror
\begin{algorithm}[H]
\LinesNumbered
\SetAlgoLined
\KwData{Pruned model parameters at this iteration $\theta$, the number of fine-tuning iterations $\mathcal{Z}$, learning rate $\gamma$ and decay of learning rate $\tau$.}
\KwResult{Further compressed and tuned model parameters $\theta$.}
\For{$\mathcal{Z}$ iterations}{
Randomly choose a mini-batch of samples from the training set\;
 Compute gradient of $\mathcal{L}(\theta; \mathbf{x}^{(i)})$ by $\frac{\partial  \mathcal{L}(\theta; \mathbf{x}^{(i)})}{\partial  \theta}$, where $\mathcal{L}(\theta; \mathbf{x}^{(i)})$ is computed by Equation~\eqref{vad}\;
 Update $\theta$ using $\theta \leftarrow \theta -  \gamma \frac{\partial  \mathcal{L}(\theta; \mathbf{x}^{(i)})}{\partial  \theta}$\;
 $\gamma \leftarrow \tau \gamma$
}
 \caption{Variational pruning.}\label{alg:2}
 \end{algorithm}
 \endgroup
  \end{minipage} }

\section{Quantization}
In the proposed DRL-based quantization-aware training, the RL agent automatically searches for the optimal bit allocation representation strategy for each layer. The modeling of quantization state, action, and rewards are defined in Section~\ref{sec:drl}. The DRL structure is the same as the one for pruning in subection~\ref{sec:DRLstructure}. In the fine-tuning step of the quantized CNN, we apply Straight-Through Estimator (STE)~\cite{bengio2013estimating}. The idea of this estimator of the expected gradient through stochastic neurons is simply to back-propagate through the hard threshold function, e.g., sigmoidal non-linearity function~\cite{yin2003flexible}. The gradient is $1$ if the argument is positive and $0$ otherwise.

\section{Experiments}\label{sec:vpe1}  
\subsection{Settings}
In all experiments, the MNIST and CIFAR-10 dataset are both divided by 50000 samples for training, 5000 samples for validation, and 5000 samples for evaluation. The ILSVRC-12 dataset is divided by 1281167 samples for training, 10000 samples for validation, and 50000 samples for evaluation. We adopt a neural network policy with one hidden layer of size 64 and one fully-connected layer using sigmoid as the activation function. We use the proximal policy optimization clipping algorithm with
c = 0.2 as the optimizer. The critic also has one hidden layer of size 64. The discounting factor is selected as $\gamma= 0.99$. The learning rate of the actor and the critic is set as $1 \times 10^{-3}$. In CIFAR-10, the per GPU batch size for training is 128 and the batch size for evaluation is 100. In ILSVRC-12, the per GPU batch size for training is 64 and the batch size for evaluation is 100. The fine tuning steps for each layer are selected as $2000$ in the quantization. The parameters are optimized using the SGD with momentum algorithm~\cite{sutskever2013importance}. For MNIST and CIFAR-10, the initial learning rate is set as 0.1 for LeNet, ResNet, and VGGNet. For ILSVRC-12, the initial learning rate is set as 1 and divided by 10 at rollouts 30, 60, 80, and 90. The decay of learning rate is set to 0.99. All experiments were performed using TensorFlow, allowing for automatic differentiation through the gradient updates~\cite{abadi2016tensorflow}, on 8 NVIDIA Tesla K80 GPUs.

\subsection{MNIST and CIFAR-10}
\begin{wraptable}{r}{7.5cm}
\caption{Results on MNIST and CIFAR-10 dataset.}\label{table:1}
\centering
\scalebox{0.7}{
 \begin{tabular}{c c c c c} 
 \hline
LeNet & Error (\%) & Para. & Pruned Para. (\%) & FLOPs (\%) \\ 
 \hline
LeNet-5 (DropPruning) & 0.73 & 60K & 87.0 & - \\ 
LeNet-5 & 0.34 & 5.94K & 90.1& 16.4 \\ 
 \hline
CIFAR-10 & Error (\%) & Para. & Pruned Para. (\%) & FLOPs (\%) \\ 
 \hline
VGGNet(Baseline) & 6.54 & 20.04M & 0 & 100 \\ 
VGGNet & 6.33 & 2.20M & 89.0& 48.7 \\ 
VGGNet  & 6.20 & 2.29M & 88.6 &49.1 \\ 
 \hline
ResNet-152 (Baseline) & 5.37 & 1.70M & 0 & 100 \\ 
ResNet-152 & 5.19 & 1.30M & 23.5& 71.2 \\ 
ResNet-152   & 5.33 & 1.02M & 40.0 &55.1 \\ 
\hline
\end{tabular}}
\end{wraptable}
The MNIST~\cite{cohen2017emnist} and CIFAR-10 dataset~\cite{krizhevsky2014cifar} consists of images with a $32\times 32$ resolution. Table~\ref{table:1} shows the performance of the proposed method. It can be observed that the proposed method can not only reduce model size but also improve the accuracy (i.e., reduce error rate). In MNIST, comparing with the most recent DropPruning method, our method for LeNet obtains $10 \times$ model compression with a slightly accuracy increase (0.68\%). In CIFAR-10, comparing with the baseline model, our method for the VGGNet achieves $9 \times$ model compression with a slightly accuracy increase (0.34\%). In addition, we compare our algorithm with the commonly adopted weight magnitude channel selection strategy and channel pruning strategy to demonstrate the importance of variational pruning. Please refer to the subsection~\ref{sec:ivp} for more details.

\subsection{ImageNet}\label{sec:ivp}
\begin{table*}[!ht]
\caption{Results on ImageNet dataset for ResNet-18 and ResNet-50 with different speed-ups.}\label{table:2}
\scalebox{0.75}{
\noindent\makebox[1.3\textwidth]{
 \begin{tabular}{c c c c c} 
 \hline
\multirow{2}{*}{Model} & \multirow{2}{*}{Top-1/Top-5 Error (\%) }& \multirow{2}{*}{Pruned Para. (\%)} & FLOPs (\%) & Speed-up~$\times$ \\ 
& & & Pruning & Pruning + Quantization\\
\hline
ResNet-18 (Baseline) & 29.36/10.02 & 0 & 100 & 1 \\ 
ResNet-18 & 30.29/10.43 & 30.2 & 71.4 & 11.4 \\ 
ResNet-18   & 30.65/11.93 & 51.0 & 44.2 &16.0 \\ 
ResNet-18   & 33.40/13.37 & 76.7 & 29.5 &28.2 \\ 
 \hline
ResNet-50 (Baseline) & 24.87/6.95 & 0 & 100 & 1 \\ 
ResNet-50 & 23.42/6.93 & 31.2 & 66.7& 12.0 \\ 
ResNet-50   & 24.21/7.65 & 52.1 & 47.6 &16.0 \\ 
ResNet-50   & 28.73/8.37 & 75.3 & 27.0 &29.6 \\ 
\hline
\end{tabular}}}
\end{table*}

\begin{figure}  
\centering
\subfloat[ResNet-18 on ILSVRC-2012.]{\includegraphics[width=6.cm,height=3.cm]{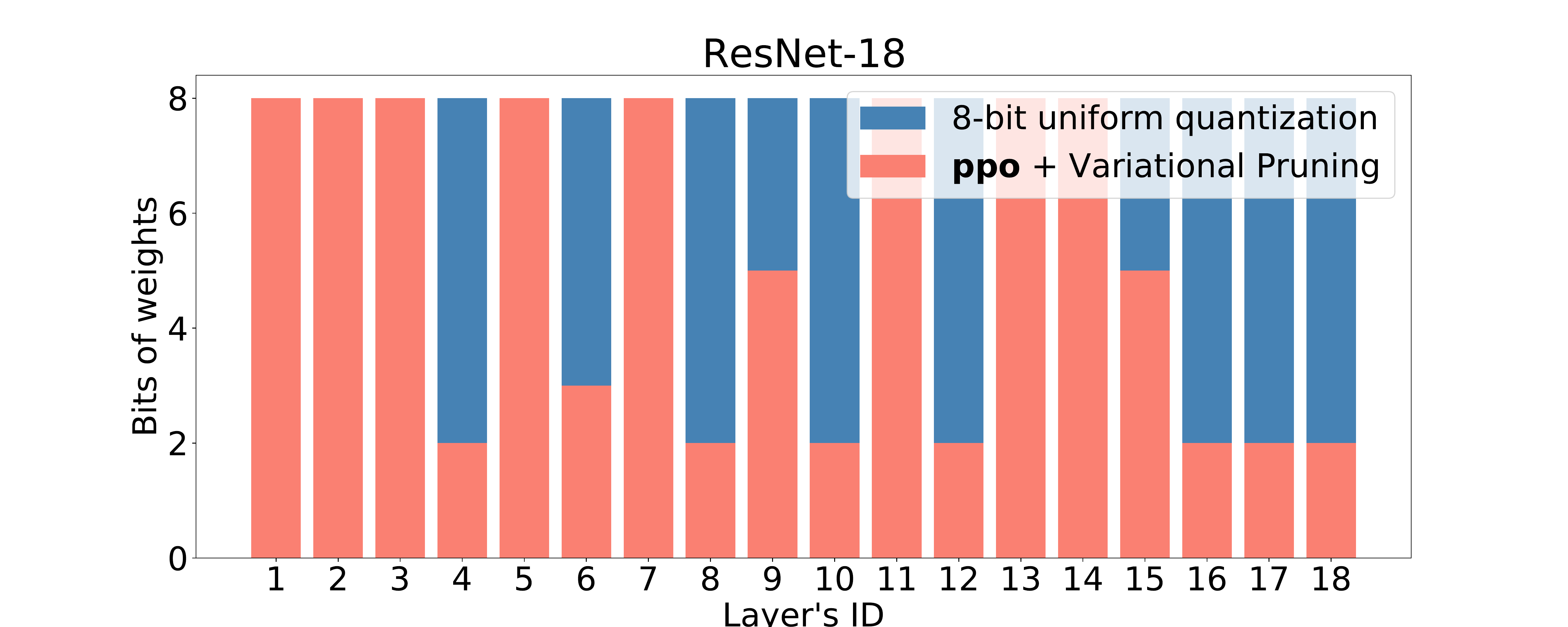}\label{1rr1}}
\subfloat[ResNet-50 on ILSVRC-2012.]{\includegraphics[width=6.cm,height=3.cm]{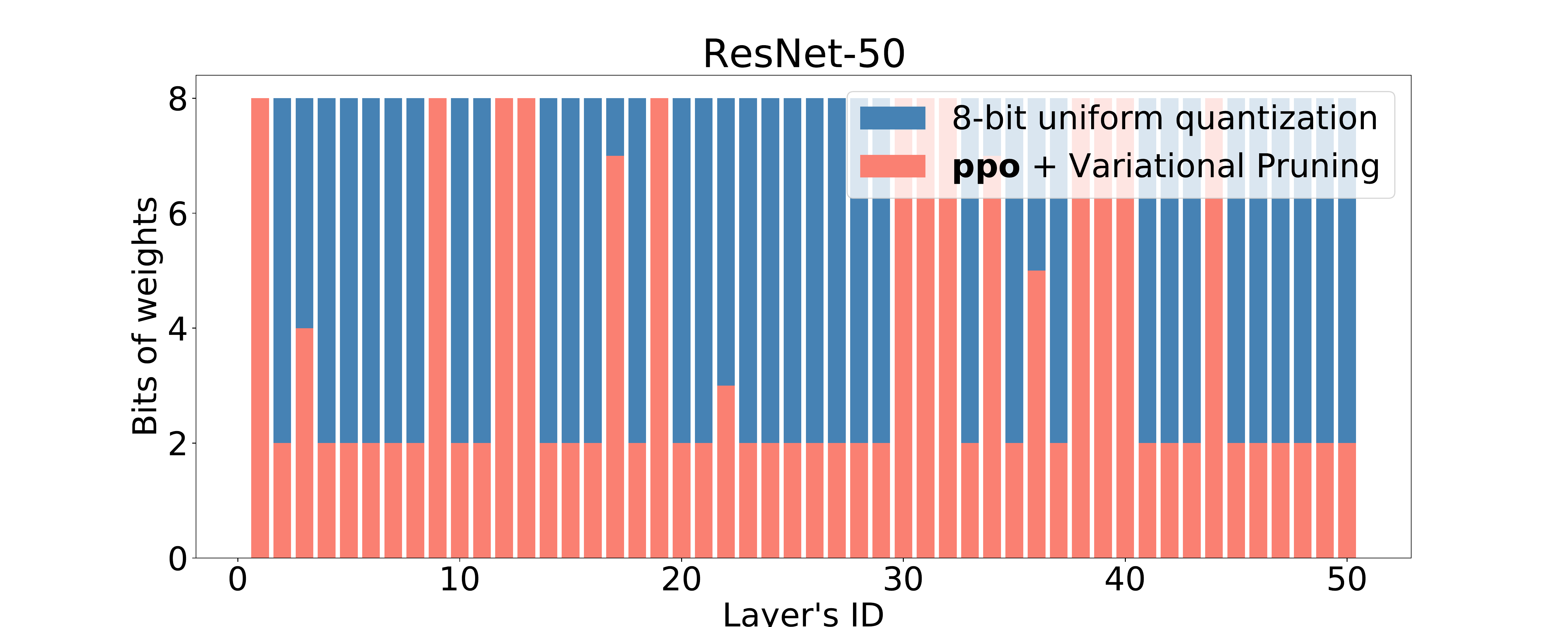}\label{1rr2}}
\caption{ResNet-18 and ResNet-50 with different
bit allocation strategies.}
\label{fig4}
\end{figure}
To evaluate the effect of different pruning rates $A_H^t$, we select 30\%, 50\%, and 70\% for ResNet-18 and ResNet-50 and then evaluate the model pruning on ImageNet ILSVRC-2012
dataset~\cite{russakovsky2015imagenet}. Experimental results are shown in Table~\ref{table:2} while the per-layer weight bits policy for the quantization is shown in Figure~\ref{fig4}. From Table~\ref{table:2}, it can be seen that the error increases as the pruning rate increases. However, our pruned ResNet-50 with 30\% pruning rate outperforms the pre-trained baseline model in the top-1 accuracy and our pruned ResNet-50 with 30\% and 50\% pruning rate outperforms the pre-trained baseline model in the top-1 accuracy. In Figure~\ref{fig4}, the 8-bit uniform quantization strategy is shown in blue bar, and the \textbf{ppo} with variational pruning policy is shown in red bar. The DRL-supported policy generates a more compressed model with a faster inference speed. By observing the DRL-supported policy, the $3\times3$ layer is more important than the $1\times1$ layers because the $1\times1$ layers are represented by less bits naturally.

\begin{wraptable}{r}{6.8cm}
\caption{MobileNet-v1 and MobileNet-v2 on ILSVRC-12.}\label{table:3}
\begin{center}
\scalebox{0.7}{
 \begin{tabular}{c c c c c} 
 \hline
Model & FLOPs (\%) & $\Delta acc (\%)$ \\ 
 \hline
MobileNet-v1 (Baseline) & 100 & 0 \\ 
 \hline
MobileNet-v1 (\textbf{ppo} + Channel Pruning) & 50 & -0.2 \\ 
MobileNet-v1 (\textbf{ppo} + Channel Pruning) & 40 & -1.1 \\ 
\hline
MobileNet-v1 (\textbf{ppo} + Variational Pruning) & 47 & +0.1 \\ 
MobileNet-v1 (\textbf{ppo} + Variational Pruning) & 40 & -0.8 \\ 
\hline
MobileNet-v1 (\textbf{ddpg})~\cite{he2018amc} & 50 & -0.4 \\ 
MobileNet-v1 (\textbf{ddpg})~\cite{he2018amc} & 40 & -1.7 \\ 
\hline
0.75 MobileNet-v1 (Uniform)~\cite{howard2017mobilenets} & 56 & -2.5 \\ 
0.75 MobileNet-v1 (Uniform)~\cite{howard2017mobilenets}  & 41 & -3.7 \\ 
\hline
\textbf{Model/Pruning} & FLOPs (\%) & $\Delta acc (\%)$ \\ 
 \hline
MobileNet-v2 (Baseline) & 100 & 0 \\ 
\hline
MobileNet-v2 (\textbf{ppo} + Variational Pruning) & 21 & -2.4 \\ 
MobileNet-v2 (\textbf{ppo} + Variational Pruning) & 59 & -1.0 \\ 
MobileNet-v2 (\textbf{ppo} + Variational Pruning) & 70 & -0.8 \\ 
\hline
MobileNet-v2 (\textbf{ddpg})~\cite{he2018amc} & 30 & -3.1 \\ 
MobileNet-v2 (\textbf{ddpg})~\cite{he2018amc} & 60 & -2.1 \\ 
MobileNet-v2 (\textbf{ddpg})~\cite{he2018amc} & 70 & -1.0 \\ 
\hline
0.75 MobileNet-v2 (Uniform)~\cite{howard2017mobilenets}  & 70 & -2.0 \\ 
\hline
\hline
\textbf{Model/Quantization} & Model Size & $\Delta acc (\%)$ \\ 
MobileNet-v2 (\textbf{ppo} + Variational Pruning) & 0.95M & -2.9 \\ 
MobileNet-v2 (\textbf{ppo} + Variational Pruning) & 0.89M & -3.2 \\ 
MobileNet-v2 (\textbf{ppo} + Variational Pruning) & 0.81M & -3.6 \\ 
MobileNet-v2 (HAQ)~\cite{wang2019haq} &0.95M & -3.3 \\ 
MobileNet-v2 (Deep Compression)~\cite{han2015deep} &0.96M & -11.9 \\ 
 \hline
\end{tabular}}
\end{center}
\end{wraptable}

To show the importance of our DRL-supported compression structure with variational pruning, we compare RL with channel pruning and RL with variational pruning. Table~\ref{table:3} shows that \textbf{ppo} with channel pruning can find the optimal layer-wise pruning rates while \textbf{ppo} with variational pruning can further decrease the testing error of the compressed model. Another observation is with the same compression scope, e.g,, $50\%$ FLOPs, our model's accuracy outperforms \textbf{ddpg} based algorithms. A comparison of the reward $r1$ for AMC~\cite{he2018amc} (DDPG-based pruning) and our proposed method (PPO-based pruning) is also shown in Figure~\ref{fig:graph} for $A_H^t = 50\%$ in $6$ runs. A typical failure mode of \textbf{ddpg} training is the Q-value overestimation, which leads to a lower reward in $r1$. We also report the results for ILSVRC-12 on MobileNet-v2 on Table~\ref{table:3}. Although for $70\%$ FLOPs, our model's performance is competitive to the \textbf{ddpg} approach, we achieve lower accuracy decrease on smaller models such as $30\%$ and $60\%$ FLOPs.

We further examine the results when applying both pruning and quantization on ILSVRC-12. We use the VGG-16 model with 138 million parameters as the reference model. Table~\ref{table:5} shows that VGG-16 can be compressed to 3.0\% of its original size (i.e., $33 \times$ speed-up) when weights in the convolution layers are represented with 8 bits, and fully-connected layers with 5 bits. Again, the compressed model outperforms the baseline model in both the top-1 and top-5 errors.
 \begin{table*}[!ht]
\caption{Comparison with another non-RL two-stage compression method (VGG-16 on ILSVRC-12).}\label{table:5}
\centering
\scalebox{0.7}{
\noindent\makebox[1.0\textwidth]{
 \begin{tabular}{|c| c c c c c|} 
 \hline
\multirow{2}{*}{Model (MobileNet-v1)} &\multirow{2}{*}{Layer} & \multirow{2}{*}{Parameters) }& \multirow{2}{*}{Pruned Para. (\%)} & Weight bits & Speed-up~$\times$ \\ 
& & & & Pruning + Quantization & Pruning + Quantization\\
\hline
\multirow{6}{*}{\textbf{ppo} + Variational Pruning} & conv1\_1 / conv1\_2 & 2K / 37K & 42 / 89 & 8 / 8 & 2.5/ 10.2 \\ 
 & conv2\_1 / conv2\_2 & 74K / 148K & 72 / 69 & 8 / 8 & 7.0/ 6.8\\ 
   & conv3\_1 / conv3\_2 / conv3\_3 & 295K / 590K / 590K & 50 / 76 / 58 &8 / 8 / 8& 4.6/ 10.3/5.9 \\ 
   & conv4\_1 / conv4\_2 / conv4\_3 & 1M / 2M / 2M & 68 / 88 / 76 &8 / 8 / 8& 7.6/ 9.1/7.2 \\ 
      & conv5\_1 / conv5\_2 / conv5\_3 & 1M / 2M / 2M & 70 / 76 / 69 &8 / 8 / 8& 7.1/ 8.5/7.1\\ 
   & fc\_6 / fc\_7 / fc\_8 & 103M / 17M / 4M & 96 / 96 / 77 &5 / 5 / 5& 62.5/ 66.7/14.1 \\ 
 \hline
 & Total & 138M & 93.1 &5& $33\times$\\
 \hline
\multirow{6}{*}{ Deep Compression~\cite{han2015deep}} & conv1\_1 / conv1\_2 & 2K / 37K & 42 / 78 & 5 / 5& 2.5/ 10.2 \\ 
 & conv2\_1 / conv2\_2 & 74K / 148K & 66 / 64 & 5 / 5 & 6.9/ 6.8\\ 
  & conv3\_1 / conv3\_2 / conv3\_3 & 295K / 590K / 590K & 47 / 76 / 58 &5 / 5 / 5& 4.5/ 10.2/5.9 \\ 
   & conv4\_1 / conv4\_2 / conv4\_3 & 1M / 2M / 2M & 68 / 73 / 66&5 / 5 / 5 & 7.6/ 9.2/7.1 \\ 
   & conv5\_1 / conv5\_2 / conv5\_3 & 1M / 2M / 2M & 65 / 71 / 64 &5 / 5 / 5& 7.0/ 8.6/6.8 \\ 
  & fc\_6 / fc\_7 / fc\_8 & 103M / 17M / 4M & 96 / 96 / 77 &5 / 5 / 5& 62.5/ 66.7/14.0 \\ 
   \hline
 & Total & 138M & 92.5 &5& $31\times$\\
\hline
\end{tabular}}}
\end{table*}

\subsection{Variational Pruning via Information Dropout}\label{sec:vpe}

\begin{figure*}[!ht]  
\centering
\subfloat[Input data $\mathbf{x}$.]{\includegraphics[width=2.5cm,height=2.5cm]{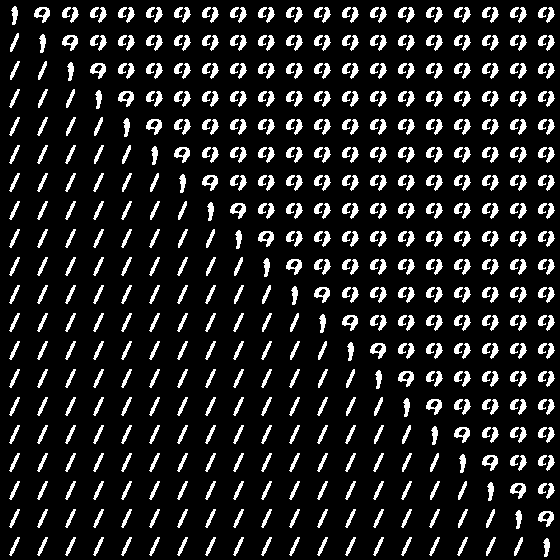}\label{rr1}}
\subfloat[Latent variable $\mathbf{z}$.]{\includegraphics[width=2.5cm,height=2.5cm]{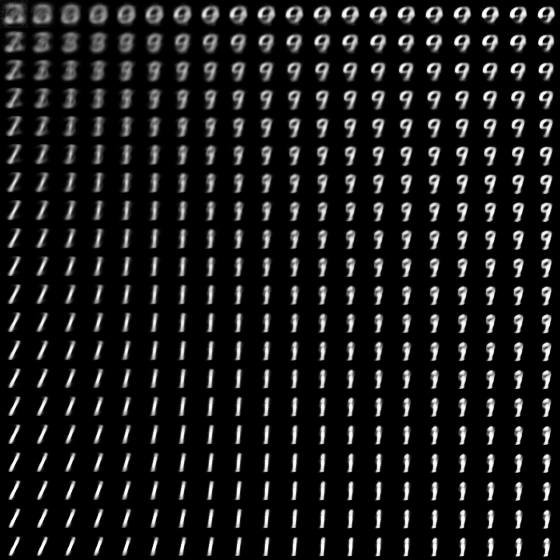}\label{rr2}}
\subfloat[Weights.]{\includegraphics[width=2.5cm,height=2.5cm]{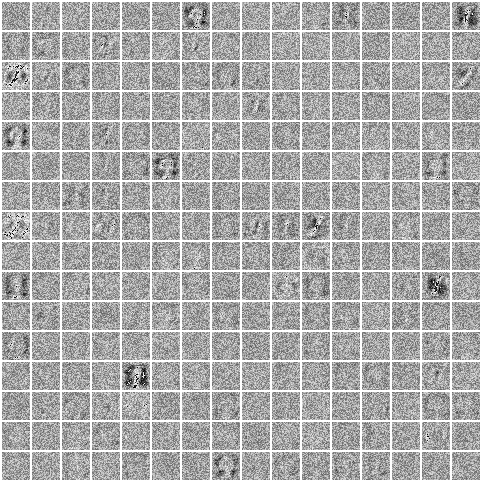}\label{rr3}}
\subfloat[The PDF of the random noise $\xi$.]{\includegraphics[width=3.0cm,height=3.0cm]{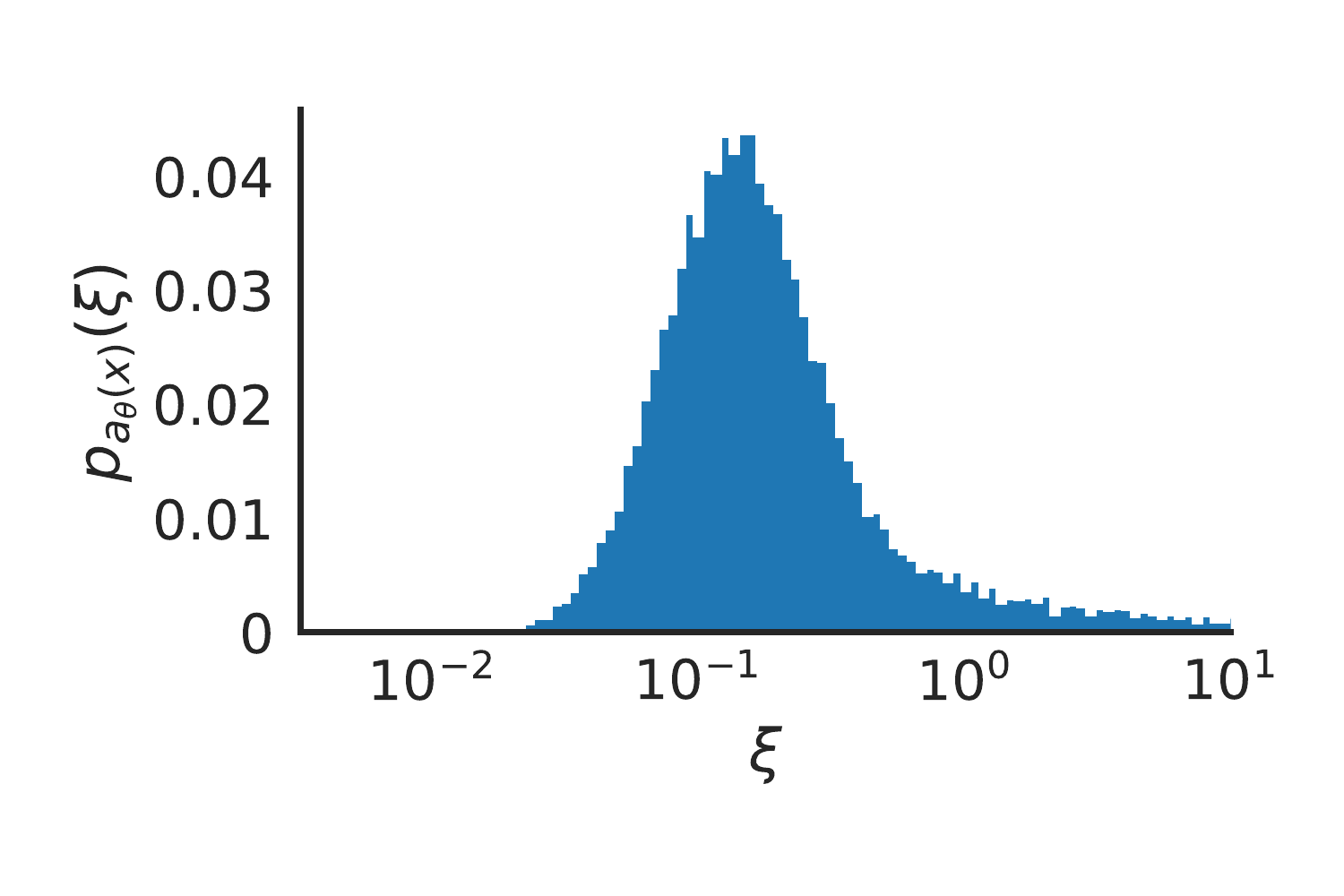}\label{rr4}}
\caption{An illustrative information dropout experiment. Figure~\ref{rr1} shows the input data $\mathbf{x}$. Figure~\ref{rr2} shows the plot of the latent variable $\mathbf{z}$ at a choice of parameter $\theta$ at each spatial location in the third information dropout layer of LeNet trained on MNIST with $\alpha = 1$. The resulting representation $\mathbf{z}$ is robust to nuisances, and provides good performance. Figure~\ref{rr3} shows the weights. Figure~\ref{rr4} shows the PDF of the noise parameter $\xi$.}
\label{fig:vae}
\end{figure*}

The goal of this illustrative experiment is to validate the approach in subsection~\ref{sec:vae11} and show that our regularized loss function $\mathcal{L}(\theta; \mathbf{x}^{(i)})$ shown in Equation~\eqref{vad} can automatically adapt to the data and can better exploit architectures for further compression. The random noise $\xi$ is drawn from a distribution $p_{a_\theta(\mathbf{x})}(\xi)$ with a unit mean $u =1$ and a variance $a_\theta(\mathbf{x})$ that depends on the input data $\mathbf{x}$. The variance $a_\theta(\mathbf{x})$ is parameterized by $\theta$. To determine the best allocation of parameter $\theta$ to minimize the KL-divergence term $D_{KL}(p_\theta(\mathbf{z}|\mathbf{x})| p_{\theta^\star}(\mathbf{z}))$, we still need to have a prior distribution $p_{\theta^\star}(\mathbf{z})$. The prior distribution is identical to the expected distribution of the activation function $f(\mathbf{x})$, which represents how much data $\mathbf{x}$ lets flow to the next layer. For a network that is implemented using the softplus activation function, a log-normal distribution is a good fit for the prior distribution (Achille and Soatto 2018). After we fix this prior distribution as $\log (p_{\theta^\star}(\mathbf{z})) \sim \mathcal{N}(0,1)$, the loss can be computed using stochastic gradient descent to back-propagate the gradient through the sampling of $\mathbf{z}$ to obtain the optimized parameter $\theta$. Even if $\log (p_{\theta^\star}(\mathbf{z})) \sim \mathcal{N}(0,1)$, the actual value of $u$ is not necessarily equal to $1$ during the runtime. Hence, the mean $u$ and the variance $a_\theta(\mathbf{x})$ of the random noise $\xi$ can be computed via solving the following two equations
 \begin{align}
 E(\xi) &= e ^{u+\frac{a^2_\theta(\mathbf{x})}{2}},\\
  D(\xi) &= (e^{a^2_\theta(\mathbf{x})}-1)e^{a^2_\theta(\mathbf{x})+2u},
 \end{align}
where $E(\xi)$ is the mean of sampled $\xi$ and $D(\xi)$ is the variance of sampled $\xi$. We add a constraint, $a_\theta(\mathbf{x}) \leq 0.8$, to avoid a large noise variance. Figure~\ref{rr4} shown the probability density function (PDF) of the noise parameter by experiment, which matches a log-normal distribution. The result shows that we optimize the parameters $\theta$ of the parameterized model $p_{\theta^\star}(\mathbf{z})$ such that $p_{\theta^\star}(\mathbf{z})$ is a close approximation of $p_{\theta}(\mathbf{z}|\mathbf{x})$ as measured by the KL divergence $D_{KL}(p_\theta(\mathbf{z}|\mathbf{x})| p_{\theta^\star}(\mathbf{z}))$. After the noise distribution is known, the distribution of $p_{\theta}(\mathbf{z}|\mathbf{x})$ in Equation~\eqref{eq:noise} can be obtained. In order to show how much information from images that information dropout is transmitting to the second layer, Figure~\ref{rr2} shows the latent variable $\mathbf{z}$ while Figure~\ref{rr3} shows the weights. As shown in Figure~\ref{rr2}, the network lets through the input data (Figure~\ref{rr1}).

\subsection{Single Layer Acceleration Performance}

In order to further show the importance of variational pruning after obtaining the optimized pruning rate based on the \textbf{ppo} algorithm, we  test a simple 4-layer convolutional neural network, including 2 convolution (conv) layers and 2 fully connected (fc) layers, for image classification on the CIFAR-10 dataset. We evaluate single layer acceleration performance using the proposed \textbf{ppo} with variational pruning algorithm in Section~\ref{sec:vae1} and compare it with the channel pruning strategy. A third typical weight magnitude pruning method is also tested for further comparison, i.e., pruning channels based on the weights' magnitude (\textbf{ppo} + Weight Magnitude Pruning).

\begin{figure}[!ht]  
\centering
\subfloat[]{\includegraphics[width=3.0cm,height=3.cm]{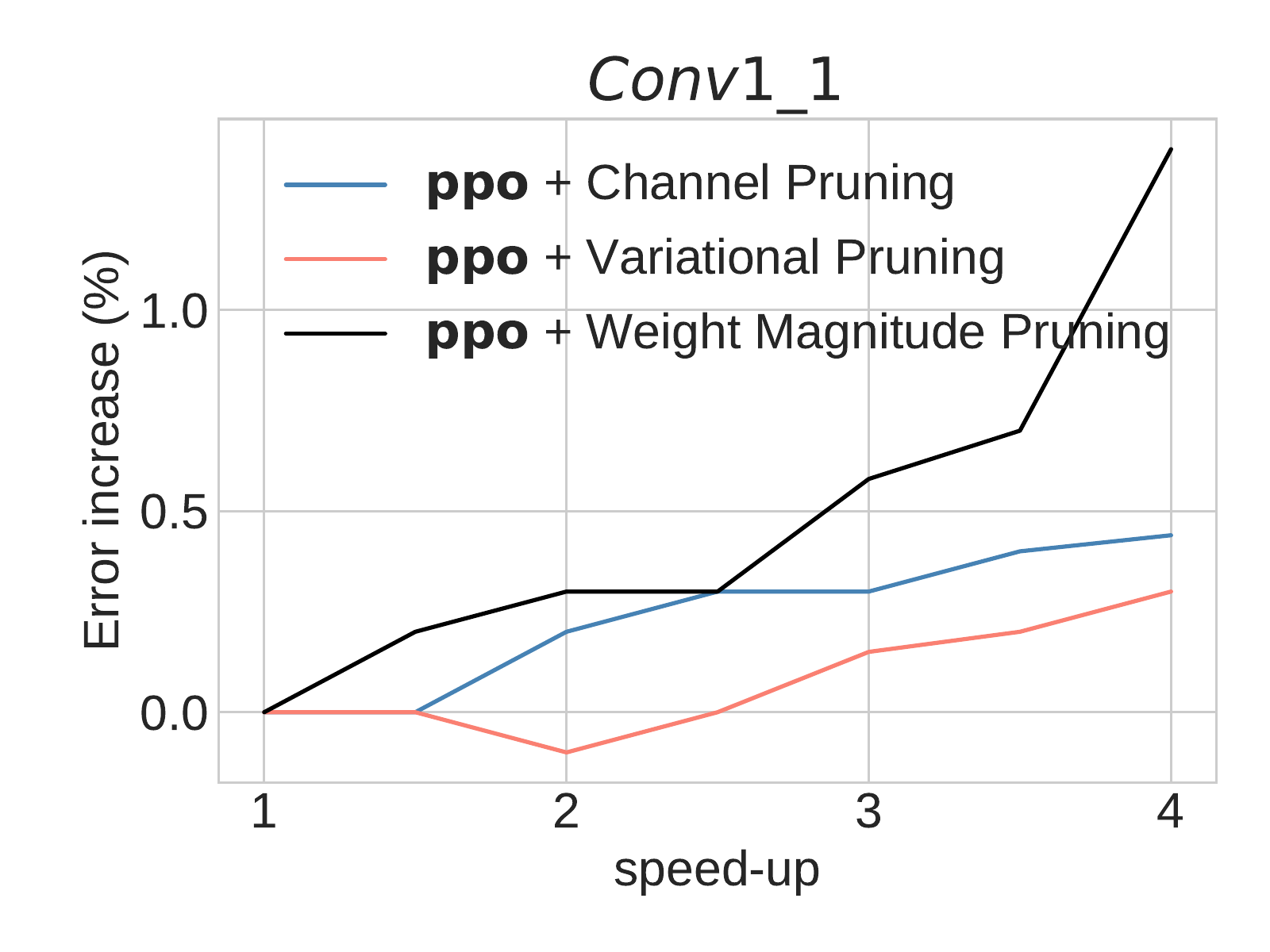}\label{11rr1}}
\subfloat[]{\includegraphics[width=3.0cm,height=3.cm]{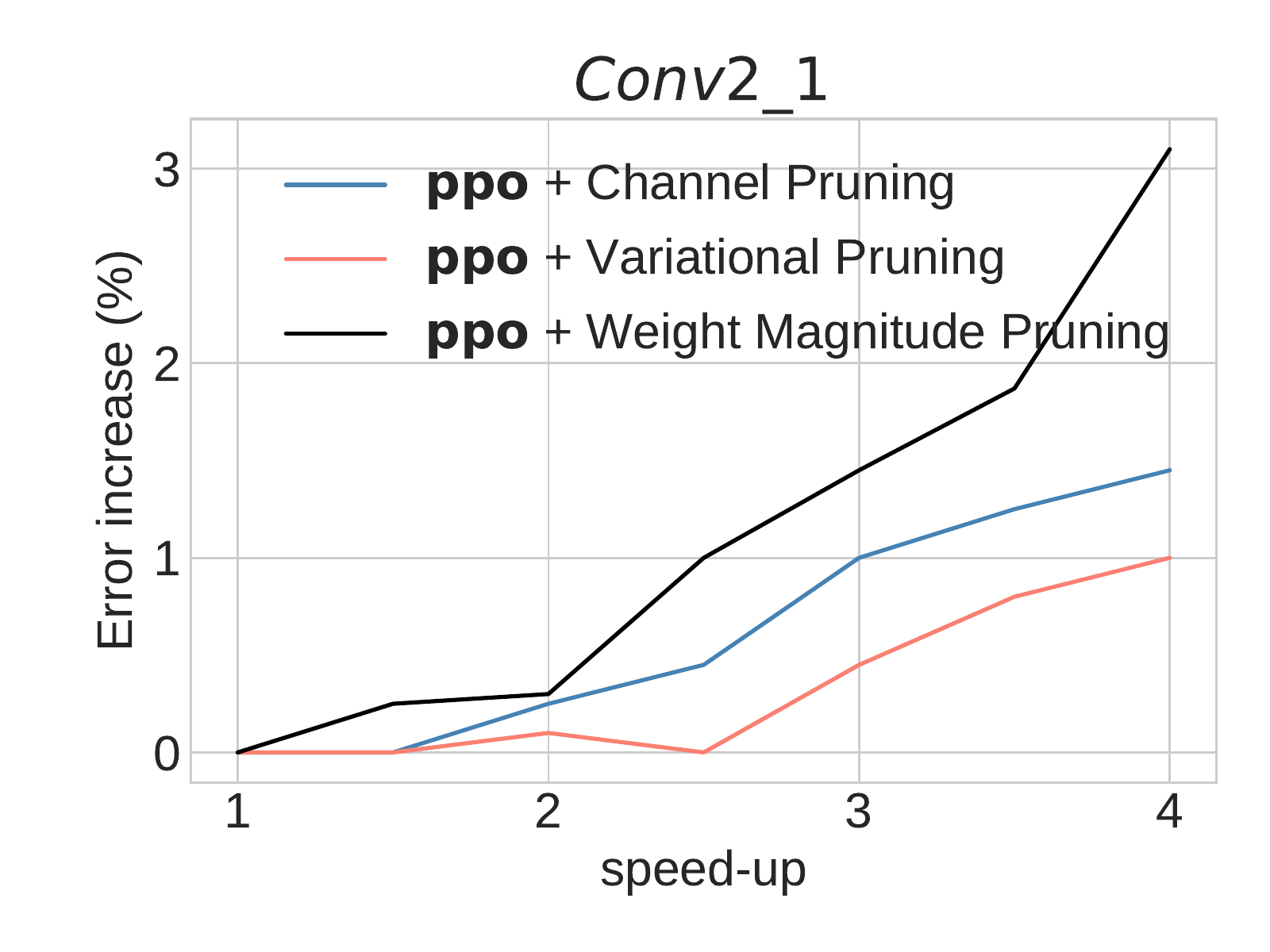}\label{11rr2}}
\subfloat[]{\includegraphics[width=3.0cm,height=3.cm]{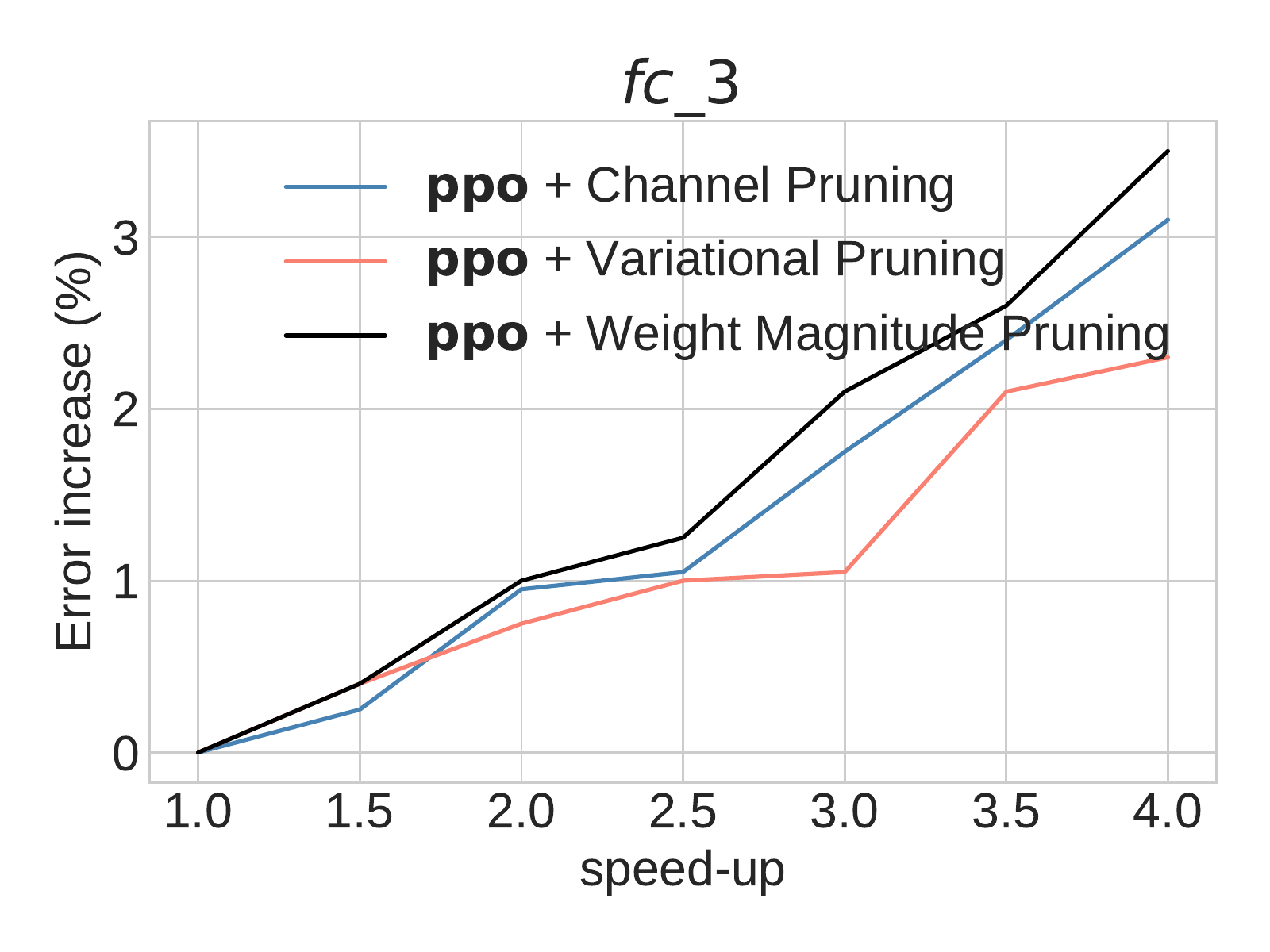}\label{11rr1}}
\subfloat[]{\includegraphics[width=3.0cm,height=3.cm]{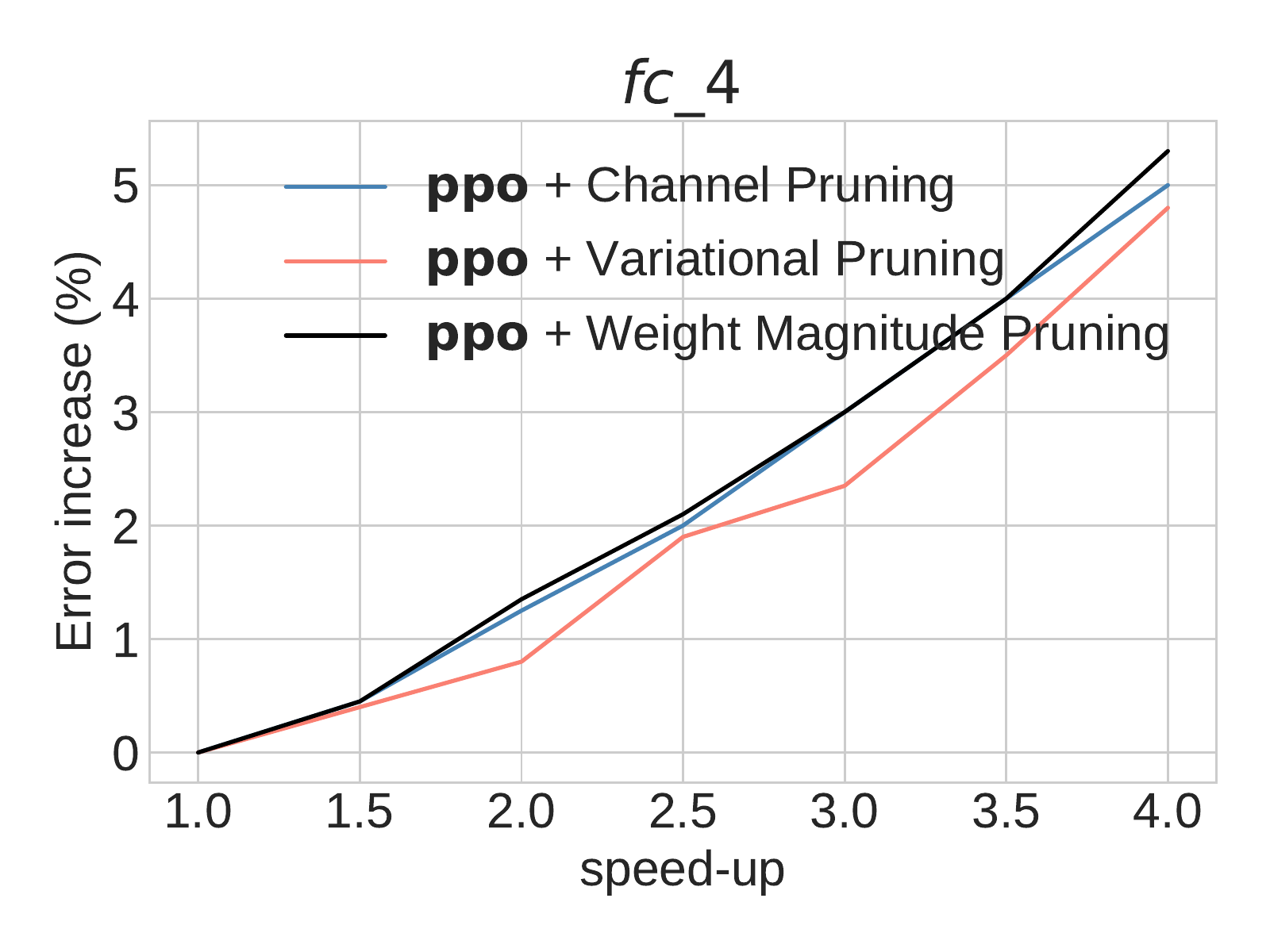}\label{11rr2}}
\caption{Single layer error increase under different compression rates. To verify the importance of variational pruning, we considered two baselines: (1) \textbf{ppo} + Channel Pruning, and (2) \textbf{ppo} + Weight Magnitude Pruning.}
\label{fig555}
\end{figure}

Figure~\ref{fig555} shows the performance comparison measured by the error
increase after a certain layer is pruned. By analyzing this figure, we
can observe that our method (\textbf{ppo} + Variational Pruning) earns the best performance in all layers. Since, \textbf{ppo} + Channel Pruning applies a LASSO regression based channel selection to minimize the reconstruction error, it achieves a better performance than the weight magnitude method. Furthermore, the proposed policy considers the fully-connected layers more important than the convolutional layers because the
error increase for fully-connected layers is typically larger under
the same compression rate.

 \subsection{Time Complexity}
 A single convolutional layer with $N$ kernels requires evaluating a total number of $NC$ of the $2D$ kernels $W_n^c \ast z^c: F = \left\{W_n^c \in \mathbb{R}^{d \times d}| n=1,\cdots,N;c=1,\cdots,C\right\}$. Note that there are $N$ kernels $F = \left\{W_n^c| n = 1,\cdots, N\right\}$ operations on each input channel $z^c$ with cost $O(NCd^2HW)$. The variational pruning via information dropout involves computing a total number of $NC^{\prime}$ of the $2D$ kernels $W_n^c \ast z^c$ with cost $O(NC^{\prime}d^2HW)$, indicating that efficiency inference requires that $C^{\prime} \ll  C$. In subsection~\ref{sec:vae11}, we consider ameliorating the inference efficiency by information dropout. In the kernels $s^c = \left\{s_m^c|m=1,\cdots,M\right\}$, the cost can be reduced to $O(NC^{\prime}d^2HW)$.

\section{Conclusion}
Using hand-crafted features to get compressed models requires
domain experts to explore a large design space and the
trade-off among model size, speed-up, and accuracy, which is
often suboptimal and time-consuming. This paper proposed
a deep model compression method that uses reinforcement
learning to automatically search the action space, improve the
model compression quality, and use the FLOPs obtained from
fine-tuning with information dropout pruning for the further
adjustment of the policy to balance the trade-off among model
size, speed-up, and accuracy. Experimental results were conducted
on CIFAR-10 and ImageNet to achieve $4\times$ - $33\times$ model compression with limited or no accuracy loss, proving
the effectiveness of the proposed method.

%

\section*{Acknowledgment}
This work was supported in part by the Army Research Office under Grant W911NF-21-1-0103.

\end{document}